\documentclass[pdflatex,sn-mathphys-num]{sn-jnl}

\usepackage{graphicx}%
\usepackage{multirow}%
\usepackage{amsmath,amssymb,amsfonts}%
\usepackage{amsthm}%
\usepackage{mathrsfs}%
\usepackage[title]{appendix}%
\usepackage{xcolor}%
\usepackage{textcomp}%
\usepackage{manyfoot}%
\usepackage{booktabs}%
\usepackage{algorithm}%
\usepackage{algorithmicx}%
\usepackage{algpseudocode}%
\usepackage{listings}%
\usepackage{enumitem}%

\usepackage{mathptmx}

\usepackage{caption}

\usepackage{xcolor} 

\usepackage{listings}
\usepackage{xcolor}

\usepackage{fancyhdr}
\theoremstyle{thmstyleone}%
\theoremstyle{thmstyletwo}%

\theoremstyle{thmstylethree}%

\begin{document}

\enlargethispage{6cm}

\title[Article Title]{Standardising the NLP Workflow: A Framework for Reproducible Linguistic Analysis}

\author[1]{\fnm{Yves} \sur{Pauli}}\email{yves.pauli@bli.uzh.ch}

\author[2]{\fnm{Jan-Bernard} \sur{Marsman}}\email{j.b.c.marsman@umcg.nl}

\author[1]{\fnm{Finn} \sur{Rabe}}\email{finn.rabe@bli.uzh.ch}

\author[1]{\fnm{Victoria} \sur{Edkins}}\email{victoria.edkins@bli.uzh.ch}

\author[1]{\fnm{Roya} \sur{Hüppi}}\email{roya.hueppi@bli.uzh.ch}

\author[2]{\fnm{Silvia} \sur{Ciampelli}}\email{s.ciampelli@umcg.nl}

\author[1]{\fnm{Akhil Ratan} \sur{Misra}}\email{akhilratan.misra@bli.uzh.ch}

\author[1]{\fnm{Nils} \sur{Lang}}\email{nils.lang@bli.uzh.ch}

\author[3]{\fnm{Wolfram} \sur{Hinzen}}\email{wolfram.hinzen@upf.edu}

\author[2]{\fnm{Iris} \sur{Sommer}}\email{i.e.c.sommer@umcg.nl}

\author*[1,4]{\fnm{Philipp} \sur{Homan}}\email{philipp.homan@bli.uzh.ch}

\affil*[1]{\orgdiv{Department of Adult Psychiatry and Psychotherapy}, \orgname{Psychiatric Hospital, University of Zurich}, \orgaddress{\street{Lenggstrasse 31}, \city{Zurich}, \postcode{8008}, \country{Switzerland}}}

\affil[2]{\orgdiv{Center for Clinical Neuroscience and Cognition}, \orgname{University of Groningen}, \orgaddress{\street{Broerstraat 5}, \city{Groningen}, \postcode{9712}, \country{Netherlands}}}

\affil[3]{\orgdiv{Department of Translation \& Language Sciences}, \orgname{University Pompeu Fabra}, \orgaddress{\street{Carrer de la Mercè, 12, Ciutat Vella}, \city{Barcelona}, \postcode{08002}, \country{Spain}}}

\affil[4]{\orgdiv{Neuroscience Center Zurich}, \orgname{University of Zurich and ETH Zurich}, \orgaddress{\street{Winterthurerstrasse 190}, \city{Zurich}, \postcode{8057}, \country{Switzerland}}}


\abstract{The introduction of large language models and other influential developments in AI-based language processing have led to an evolution in the methods available to quantitatively analyse language data. With the resultant growth of attention on language processing, significant challenges have emerged, including the lack of standardisation in organising and sharing linguistic data and the absence of standardised and reproducible processing methodologies. Striving for future standardisation, we first propose the Language Processing Data Structure (LPDS), a data structure inspired by the Brain Imaging Data Structure (BIDS), a widely adopted standard for handling neuroscience data. It provides a folder structure and file naming conventions for linguistic research. Second, we introduce pelican\_nlp, a modular and extensible Python package designed to enable streamlined language processing, from initial data cleaning and task-specific preprocessing to the extraction of sophisticated linguistic and acoustic features, such as semantic embeddings and prosodic metrics. The entire processing workflow can be specified within a single, shareable configuration file, which pelican\_nlp then executes on LPDS-formatted data. Depending on the specifications, the reproducible output can consist of preprocessed language data or standardised extraction of both linguistic and acoustic features and corresponding result aggregations. LPDS and pelican\_nlp collectively offer an end-to-end processing pipeline for linguistic data, designed to ensure methodological transparency and enhance reproducibility.}

\keywords{NLP, Reproducibility, Pipeline, Preprocessing, Linguistic Metric Extraction, Data Structure, Python package}



\maketitle

\section{Introduction}\label{Introduction}

The analysis of linguistic data is central to research in numerous disciplines, including psychiatry, psychology, neuroscience\cite{crema2022nlp_clinical} and behavioural sciences\cite{feuerriegel2025nlp_behavioral}. Advances in natural language processing (NLP) following the introduction of transformer-based language models\cite{vaswani2017attention} have revolutionised the field of language research, providing numerous exciting new opportunities. Researchers employ a variety of tools and techniques to process language data\cite{bird2009nlp_python,manning2014corenlp,honnibal2020spacy,akbik2019flair,qi2020stanza,wolf2020transformers,gardner2018allennlp,falcon2019pytorch}, including both traditional linguistic methods and various AI-driven models. Essential preprocessing, including data cleaning, normalisation, and tokenisation, prepares the data for subsequent quantitative analysis. This often involves extracting informative features from both the textual content (e.g., word embeddings\cite{mikolov2013efficientestimationwordrepresentations}) and the speech signal itself (e.g., acoustic features, to make linguistic phenomena computationally tractable. However, a lack of standardisation in language processing pipelines, including data storage, data preprocessing and metric extraction, has led to a lack of reproducibility and comparability between studies\cite{fokkens2013offspring,belz2022metrological_repro_lab,mieskes2019community_replication_ranlp,belz2024repronlp_shared_task_2024,howcroft2020twenty_years_confusion_nlg}. 

This lack of standardised formats for storing and sharing linguistic data frequently results in the need to reformat datasets to accommodate different tools and methodologies. This redundancy persists even when processing steps are similar, resulting in inefficiency and hampering collaborations\cite{article,wilkinson2016fair}. Additionally, language processing, involves numerous researcher degrees of freedom, for example, relating to choices in implementation and application of various preprocessing steps, including lemmatisation, speaker diarisation, handling missing data, or including special character tokens. Researcher degrees of freedom undermine reproducibility\cite{munafo2017manifesto,baker2016reproducibility_survey,ioannidis2005why_most_false,wieling2018repro_computational_linguistics,mieskes2017quantitative_acl_ethics,belz2023missing_info_negative_results_nlp} and have been shown to significantly affect research outcomes both in NLP research\cite{fokkens2013offspring,denny2018text_preprocessing,schofield2016stemmers_topic_models,crane2018questionable_qa,belz2021reprogen_shared_task_2021,branco2020reprolang2020,belz2024repronlp_shared_task_2024,belz2021systematic_review_reproducibility_nlp,belz2022repronlp_reprogen2022, belz2023repronlp_shared_task_2023}, as well as other areas\cite{botvinik2020variability,silberzahn2018many_analysts,carp2012analytic_flexibility,wicherts2016degrees_freedom,simmons2011false_positive_psychology}, to the point where contrary results have been reported based on the analysis of the same dataset\cite{schweinsberg2021same_data,doi:10.1073/pnas.2203150119}. To enable reproducibility of language research, tools and methodologies are needed that allow for homogeneous data processing and consistent extraction of linguistic features.

To address these challenges, we introduce the Language Processing Data Structure (LPDS), a standard for organising language data, and pelican\_nlp (Preprocessing and Extraction of Linguistic Information for Computational Analysis – Natural Language Processing), a language processing Python package. By establishing LPDS and pelican\_nlp as complementary solutions, we provide a comprehensive framework for linguistic data processing and sharing. These tools aim to promote reproducibility in the field, facilitating transparency, homogenisation, and quantitative comparisons across linguistic studies by enabling the creation of standardised language processing pipelines. 

\subsection{Development of LPDS and pelican\_nlp }\label{Development}

The design of LPDS was inspired by the Brain Imaging Data Structure (BIDS)\cite{gorgolewski2016bids}, a standard that has successfully fostered collaboration and enabled robust tool development within neuroimaging\cite{poldrack2017repro_neuroimaging}. Key principles adopted for this project include a predefined hierarchical directory structure reflecting experimental design and descriptive file naming using controlled key-value pairs (entities). This approach renders data organisation both human-readable and machine-actionable, critically enabling automated discovery and processing while ensuring rich metadata description crucial for linguistic data (e.g., task type, acquisition parameters). 

The development of LPDS and pelican\_nlp was iterative and directly driven by research requirements encountered in the field of computational psychiatry. Initial specifications for LPDS and core pelican\_nlp modules for word embedding extraction from clinical interviews were developed and validated during uncertainty modelling in multimodal speech analysis\cite{rohanian2025uncertaintymodelingmultimodalspeech}. Subsequent work connected to the TRUSTING project funded by the European Union's Horizon Europe research and innovation programme\cite{trusting_about, huppi2025trusting} on analysing verbal fluency tasks and artificially created language data spurred several key developments. To support these new analyses, the use of an additional language model was integrated into the pipeline\cite{bojanowski2017subword}, modules for semantic similarity were implemented, and the LPDS standard was generalised to accommodate the respective data formats.
To address the requirements of a further project, connected to TRUSTING, on speech-based prediction in psychosis, modules for acoustic feature extraction using the openSMILE\cite{Eyben2010openSMILE} and prosogram\cite{Mertens2004Prosogram} toolkits were added to the framework. This practice-driven development informed the specific design choices detailed below.

Complementing the data standard, pelican\_nlp was developed as a modular Python package to standardise common language processing steps. Similar to already established, robust pipelines like fMRIprep in neuroimaging\cite{esteban2019fmri_prep}, pelican\_nlp aims to encapsulate complex or variable procedures into a single, reproducible workflow\cite{peng2011reproducible_computational_science,sandve2013ten_rules}. The core functionalities implemented in pelican\_nlp span three primary categories: configurable preprocessing, standardised semantic feature extraction and standardised acoustic feature extraction. Preprocessing modules address common sources of variability, currently including routines for data cleaning (handling timestamps, special characters, whitespaces, case conversion, punctuation removal), speaker identification for discourse analysis, and specialised processing for tasks like verbal fluency. Concurrently, the package provides standardised extraction capabilities for key linguistic features such as semantic embeddings, semantic similarity measures, and model logits, as well as established acoustic feature sets. A summarising framework of the pelican\_nlp package is shown in Figure 1.

Several key design decisions were made to maximise the accessibility of pelican\_nlp and encourage its widespread adoption within the research community. The choice of Python as the core implementation language ensures broad accessibility due to its widespread adoption and extensive libraries. Furthermore, the use of a detailed YAML configuration file allows researchers to precisely specify and share their entire processing pipeline without requiring extensive programming expertise. This detailed pipeline specification explicitly documents analytical choices and therefore directly facilitates replication efforts\cite{barnes2010publish_code}. Consequently, the application of complex, standardised analyses becomes easily achievable. The primary user interaction involves adapting parameters within the well-documented configuration file to define the desired processing pipeline. This is followed by execution of the pipeline with a simple command-line call (‘pelican-run’), minimising the need for extensive custom programming or scripting.

\begin{figure}[htbp]
    \centering
    \fcolorbox{black}{white}{\includegraphics[width=1\textwidth]{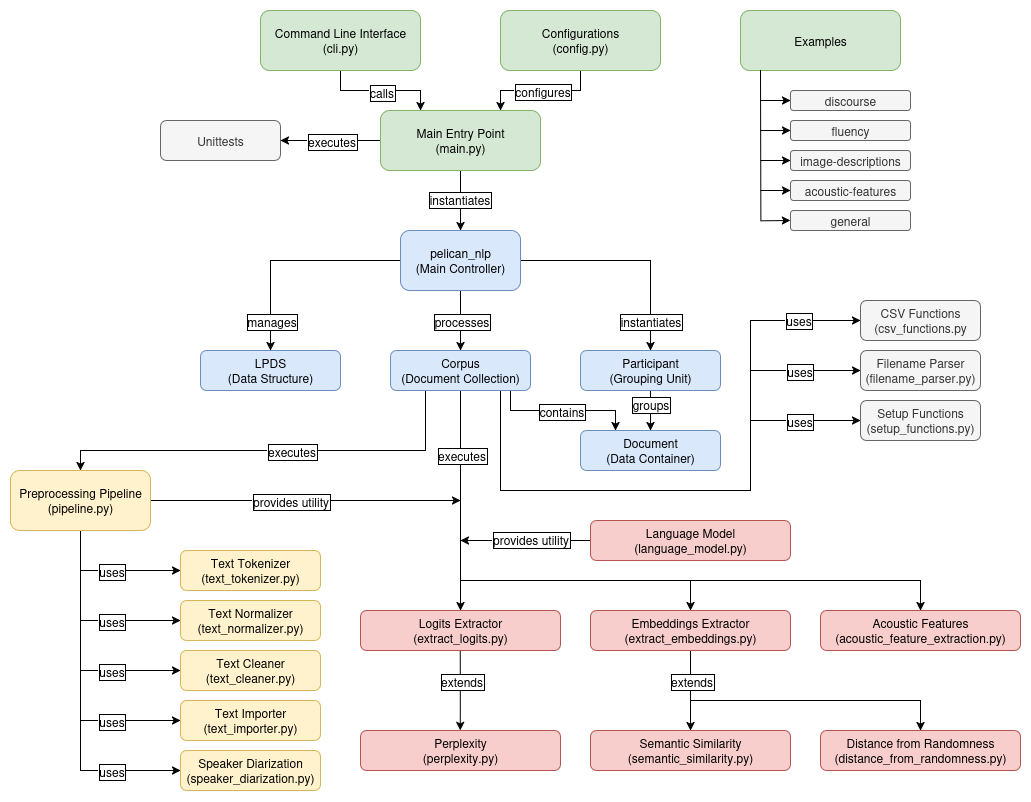}}
    \caption{Framework of the pelican\_nlp package. The framework shows processing details from command line interface to linguistic feature output. Green boxes represent core processing files. Blue boxes correspond to the main components of the package. Yellow boxes correspond to files related to data preprocessing. Red boxes correspond to files related to linguistic feature extraction. Grey boxes correspond to package utility files.}
    \label{fig:pelican_graph}
\end{figure}

\subsection{Applications of LPDS and pelican\_nlp }\label{Application}

LPDS is a proposed standardised structure for storing any type of linguistic data. This includes both audio and text files containing diverse linguistic content, such as interview recordings and their transcripts, speech corpora, or data derived from various experimental speech tasks like verbal fluency assessments or image descriptions. While our goal for LPDS is to eventually provide guidelines for all types and structures of linguistic data, we recognise that achieving universal coverage immediately is challenging. For this reason, LPDS will be iteratively adapted to accommodate incompatible edge cases when possible. 

This protocol, combining the LPDS data structure and the pelican\_nlp processing pipeline, is designed to empower researchers across various disciplines (including clinical neuroscience, psychology, behavioural sciences, and linguistics) to conduct standardised and reproducible quantitative analyses of language data. It lowers the technical barrier for implementing complex analysis pipelines, making sophisticated multimodal feature extraction accessible. This includes semantic analysis (e.g., embedding analysis, similarity computations) as well as acoustic feature processing (e.g., extracting prosody, vocal jitter, or shimmer from toolkits like openSMILE), without requiring users to manually integrate numerous disparate tools or possess extensive programming expertise for every step. Crucially, this framework directly addresses the issue of analytical variability, where different research teams have reported contrary findings from the same dataset\cite{schweinsberg2021same_data,doi:10.1073/pnas.2203150119}. By capturing the entire processing pipeline, from preprocessing choices to feature extraction parameters, within a single, version-controlled configuration file, our protocol provides an explicit and shareable blueprint for an analysis, thereby eliminating the undocumented researcher degrees of freedom that lead to divergent results. Consequently, the framework is particularly well-suited for a range of research applications, including comparative group studies (e.g., clinical vs. control), longitudinal research tracking linguistic change over time or in response to interventions, multi-site investigations requiring harmonised processing across different data acquisition locations, biomarker identification from speech or text, and computational studies focused on replicating or extending previous findings.

By using LPDS and versioning the pelican\_nlp package, the framework ensures that analyses can be faithfully reproduced even years after their original execution. Researchers who specify the package version used in a given study can rerun the exact same workflow on the same data, guaranteeing compatibility and preserving reproducibility in line with FAIR (findable, interoperable, accessible, and reusable) principles\cite{wilkinson2016fair}. This backward compatibility not only supports immediate applications such as multi-site harmonisation and longitudinal tracking, but also safeguards future efforts to replicate or extend prior findings without the risk of broken dependencies or changing outputs over time.

A key aspect of the framework's versatility is its support for multiple languages. The language-agnostic design of pelican\_nlp means its text processing capabilities are primarily determined by the underlying language models specified in the configuration file. The use-cases in this protocol leverage models such as RoBERTa\cite{liu2019roberta}, Llama3\cite{touvron2023llama}, and fastText\cite{Joulin2016fastText}, thus guaranteeing robust support for the wide range of languages covered by these architectures. Similarly, the acoustic feature extraction is designed to be broadly applicable. While the acoustic analysis modules, which integrate functionalities from toolkits like Praat\cite{BoersmaWeeninkPraat}, were validated using Dutch language data for this work, the extracted features themselves (e.g., pitch, intensity, jitter, shimmer) are fundamental properties of the speech signal. As such, the acoustic analysis pipeline is not limited to a specific language and can be applied to audio data from any spoken language.

\subsection{Comparison with other methods}\label{Comparison}

The field of NLP benefits from a rich ecosystem of powerful toolkits and libraries, each with distinct strengths. However, they are often specialised components rather than end-to-end solutions for reproducible linguistic research. Broadly, they can be grouped as follows: \\

\textbf{General-purpose educational toolkits} \\
The Natural Language Toolkit (NLTK) provides hundreds of algorithms and data structures for tokenisation, part-of-speech tagging, syntactic parsing, and basic semantic analysis\cite{bird2009nlp_python}. Its strength lies in its modularity and educational value, making it ideal for teaching core concepts and prototyping. However, this modularity comes at the cost of integration, as users must manually connect individual components into a cohesive workflow. Critically, the toolkit neither enforces a standardised data structure for interoperability nor provides optimisation for high-throughput, reproducible analysis. Addressing these specific gaps is the primary motivation for the LPDS and pelican\_nlp framework. \\

\textbf{High-performance text-annotation libraries} \\
Libraries like spaCy\cite{honnibal2020spacy} and Stanza\cite{qi2020stanza} offer production-grade, optimised pipelines for core text-annotation tasks such as named-entity recognition and dependency parsing. They excel at processing large volumes of clean, written text with high speed and accuracy. The key distinction lies in the scope of the workflow: spaCy and Stanza are specialised text-annotation engines that assume pre-processed text as input, whereas pelican\_nlp manages a broader, end-to-end pipeline that includes initial data ingestion (from both audio and text), LPDS-based organisation, standardised preprocessing, and both semantic and acoustic feature extraction. \\

\textbf{Deep linguistic analysis libraries} \\
A number of frameworks offer deep, linguistically-informed analyses. A prominent example is Stanford CoreNLP\cite{manning2014corenlp}, a comprehensive, Java-based framework with a rich set of annotators for advanced tasks like coreference resolution and sentiment analysis. While its analyses are linguistically deep, it presents two gaps that our protocol addresses: its Java-centric design creates integration challenges in Python-dominant research environments, and more importantly, it lacks native support for an end-to-end workflow that includes raw audio processing or a standardised data schema like LPDS for ensuring interoperability. \\

\textbf{Scalable topic modelling and semantic analysis libraries} \\
A distinct class of libraries focuses on unsupervised topic modelling and semantic analysis at scale. A prominent example is Gensim, an open-source Python library specialised in unsupervised topic modelling (e.g., Latent Dirichlet allocation (LDA)) and efficient training of word embeddings from large corpora. The core advantage of such tools is scalability, allowing it to process datasets that exceed available memory. The purpose of pelican\_nlp is complementary: rather than being a specialised tool for a specific analysis like topic modelling, it is a higher-level framework for standardising the entire preprocessing and feature extraction workflow, within which models for generating embeddings (similar to those Gensim can train) are applied. \\

\textbf{Contextual embedding toolkits} \\
Several libraries are exceptional at their specific function: providing access to state-of-the-art models and inference capabilities. For example, Flair\cite{akbik2019flair} simplifies the application of contextual embeddings to sequence-labeling tasks, while Hugging Face Transformers\cite{wolf2020transformers} provides a unified API to thousands of pre-trained models (e.g., BERT\cite{devlin2019bert}, RoBERTa\cite{liu2019roberta}, Llama\cite{touvron2023llama}). However, they are designed to be components within a larger pipeline, not an entire pipeline itself; they operate on the assumption that input text has already been cleaned and prepared, and they do not address standardised data organisation, preprocessing, or the structured output of final linguistic features. \\

\textbf{Research-oriented modelling toolkits} \\
Frameworks like AllenNLP\cite{gardner2018allennlp} (in maintenance since 2022) and PyTorch Lightning\cite{falcon2019pytorch} provide high-level abstractions to streamline the process of designing, training, and evaluating novel neural architectures. They make experimentation easier compared to writing raw PyTorch or TensorFlow code. Their focus, however, is on facilitating flexibility for model development, whereas pelican\_nlp is designed for the standardised and reproducible application of existing models within a fixed research protocol. \\

Taken together, these toolkits provide powerful building blocks. However, a researcher aiming to construct a complete, reproducible end-to-end pipeline, from unprocessed source files, such as raw audio recordings and their initial transcripts, to final analysis, still faces several systemic challenges that these tools do not individually solve. The most significant of these gaps, which motivated the development of our protocol, are:

\begin{enumerate}
\item Absence of a unified data schema: No single community-adopted standard exists for representing transcripts, time-aligned annotations, speaker labels, and audio-derived prosodic features, which hinders data sharing and pipeline reuse.

\item Fragmented component chaining: Critical steps like speech-to-text, tokenisation, embedding, and feature calculation exist in separate packages. Chaining them requires significant effort to overcome compatibility issues and data conversions, making precise replication difficult.

\item Reproducibility and deployment gaps: Without standardised configurations and version control for the entire workflow, small differences in parameters or underlying tool versions can lead to divergent outcomes.

\item Privacy and compliance hurdles: Processing sensitive data from healthcare or legal contexts requires secure, on-premises workflows that existing cloud-based solutions may not satisfy.

\end{enumerate}

In conjunction with LPDS, pelican\_nlp is designed to specifically address these challenges. It distinguishes itself not by attempting to outperform specialised libraries at every individual task, but by integrating the necessary components into a cohesive, standardised, and reproducible workflow tailored for linguistic research. This workflow provides the following features: 

\begin{itemize}
\item Standardised input/output: Data consumed by pelican\_nlp is structured according to LPDS, and its outputs (preprocessed data, extracted features) adhere to LPDS naming conventions within a dedicated derivatives directory. The BIDS-inspired structure of LPDS provides a natural mechanism for tracking provenance by allowing for the creation of files containing additional information, such as the version of pelican\_nlp, the specific parameters from the configuration file or parameters of contributing packages, alongside the output. This capability for detailed record-keeping is fundamental to facilitating transparent replication, data management, sharing and interoperability. 

\item Integrated pipeline: pelican\_nlp offers a configurable pipeline that handles steps often requiring separate tools, including loading text/audio, text preprocessing (handling timestamps, special characters, whitespace, case, punctuation, speaker tags), specialised cleaning routines (e.g., for fluency tasks), and extraction of various linguistic and acoustic features (e.g., semantic embeddings, similarity scores, logits, and feature sets from openSMILE and Prosogram).
\item Explicit configuration: The entire processing pipeline—including preprocessing choices, model selection (where applicable), and feature extraction parameters—is defined within a single, shareable YAML configuration file. This explicit documentation of methodology reduces ambiguity and enhances reproducibility. 
\item Focus on linguistic research needs: Modules and features are chosen based on common requirements in quantitative linguistic analysis, particularly in fields like computational psychiatry and behavioural sciences. 
\end{itemize}

Therefore, pelican\_nlp provides a higher-level framework that standardises the entire workflow from structured data input through configurable processing to structured feature output. By harmonising these steps and leveraging the LPDS standard, pelican\_nlp fills a critical niche, facilitating more transparent, comparable, and reproducible computational linguistic research. 

\subsection{Limitations}\label{Limitations}

LPDS and pelican\_nlp simplify reproducible language processing by enforcing a single data format and end-to-end pipeline. However, several limitations remain, reflecting both the complexity of the field and opportunities for future extensions. These limitations can be summarised as follows: 

\begin{enumerate}
\item LPDS scope and extensibility: Although designed for extensibility, the current LPDS specification primarily targets common linguistic data types like audio recordings and text transcripts. Accommodating highly complex or multimodal data structures or integrating highly specialised annotation formats, may require further specification development and community consensus. Furthermore, retrofitting large existing datasets to the LPDS standard can demand significant initial effort. 

\item Feature completeness: While pelican\_nlp integrates numerous common preprocessing and feature extraction steps, it does not encompass every possible NLP technique or cutting-edge linguistic metric. Researchers with highly specialised needs might still need to implement custom modules or combine pelican\_nlp outputs with other specialised tools. However, as pelican\_nlp is still under active development, incorporation of new features is ongoing. Furthermore, the framework is designed to be extensible, and we welcome community contributions of new modules via pull requests, allowing the tool to evolve and adapt to a wider range of research applications.

\item Dependency on underlying models: The performance and outputs of pelican\_nlp (e.g., embeddings, logits) inherently depend on the underlying language models utilised (e.g., specific transformer models accessed via libraries like Hugging Face). Users should be aware that potential biases present in these pre-trained models may be inherited in the extracted features. Furthermore, as the field of NLP evolves rapidly, ensuring compatibility with the latest models requires ongoing package maintenance and updates. 

\item Configuration flexibility and validation: pelican\_nlp offers a high degree of configurability via the YAML file, allowing pipelines to be tailored to diverse research questions. While this flexibility is a strength, users employing novel parameter combinations should exercise standard scientific diligence in validating the appropriateness and correctness of the outputs for their specific research context and data. 

\item Performance and scalability: While optimised for Linux environments with GPU acceleration, processing extremely large datasets or using highly complex configurations may still pose computational challenges. Performance on other operating systems or hardware configurations (e.g., CPU-only, non-NVIDIA GPUs) might be limited for certain computationally intensive tasks.
\end{enumerate}

\section{Materials}\label{Materials}

\subsection{Equipment setup}

\subsubsection{Hardware}\label{Hardware}
A standard modern processor is suitable for running the protocol. This protocol was developed and tested on a 13th Gen Intel® Core™ i9-13900H × 20 processor and an NVIDIA GeForce RTX™ 4090 Laptop GPU. A minimum of 16 GB of RAM is recommended, although processing very large files or utilising large language models may require additional memory. For computationally intensive tasks, such as those involving large transformer models, access to an NVIDIA GPU with CUDA support and at least 16 GB of virtual RAM (VRAM) is highly recommended. While some tasks may run on lower-specification GPUs or on a CPU alone, this is at the user's discretion and may come with potential performance limitations. 

\subsubsection{Software}\label{Software}
The execution of this protocol primarily relies on the Python programming language, with version 3.10 or later being recommended for optimal compatibility. Central to the protocol is the pelican\_nlp Python package, which can be installed via pip within a dedicated environment manager, such as Conda, as detailed in the Procedure section \ref{pelican}. Users are advised to employ the latest stable version available from the Python Package Index (PyPI) (\url{https://pypi.org/project/pelican-nlp/}). This protocol was developed and tested using pelican\_nlp version 0.3.4. While pelican\_nlp is developed and optimised for Linux-based operating systems like Ubuntu, its core features are also available on Windows and MacOS. However, using Windows or MacOS will limit GPU acceleration and optimal performance and may result in unanticipated issues. Key dependencies for pelican\_nlp include standard scientific Python libraries (e.g., NumPy, Pandas) and core NLP/ML libraries (e.g., PyTorch, Transformers), which are typically handled automatically during the pip installation process.

\subsubsection{Data}\label{InputData}
The protocol assumes that all input data, encompassing audio and/or text files along with associated metadata, is organised according to the LPDS format. The specifics of this format are detailed in the Procedure section. Strict adherence to the LPDS folder structure and naming conventions is crucial, as this enables pelican\_nlp to automatically discover and correctly process the data. 

\subsubsection{Code Availability}\label{SoftwareAvailability}
The pelican\_nlp package is readily available for installation from PyPI at \url{https://pypi.org/project/pelican-nlp/}. Its source code, comprehensive documentation, issue tracker, and sample configuration files are publicly hosted on GitHub at \url{https://github.com/ypauli/pelican\_nlp}. As an open-source project, we encourage community engagement, including bug reports via the issue tracker and feature contributions through pull requests. Similarly, the detailed specification for the LPDS, which includes examples, can be found in the specified repository. 

\section{Procedure}\label{Procedure}

\subsection{Part 1: LPDS}\label{LPDS}

LPDS introduces guidelines for folder structures and naming conventions of linguistic data. When possible, the user should adhere to the following guidelines regarding linguistic data storage.

\subsubsection{Folder Structure}\label{FolderStructure}

The following steps describe the procedure to organise linguistic data according to the LPDS format. 

\begin{enumerate}

\item Establish the main project directory 

First, decide on a location for the main project directory. A common practice for data processing is to use the home directory. 

For example, the main project directory path might be defined as: 

\begin{lstlisting}[language=bash]
/home/username/my_project
\end{lstlisting}

\item Create core project subdirectories 

Within the main project directory create a subdirectory named participants:

\begin{lstlisting}[basicstyle=\ttfamily\small,breaklines=true]
/home/username/my_project/participants
\end{lstlisting}

Optionally, in the main project directory supplementary information about the project can be provided. This can include files such as dataset\_description.json, participants.tsv, README and CHANGES. \\

\item Organise data by participant 

Within the participants subdirectory create a separate folder for each participant in the study. 

Each participant-specific folder should be named part-X, where X is the unique identification label for that participant (e.g., part-01, part-02, etc.). 

For example, the directory for participant 01 would be: 

\begin{lstlisting}[basicstyle=\ttfamily\small,breaklines=true]
/home/username/my_project/participants/part-01
\end{lstlisting}

Within each participant’s folder include, if desired, a participant\_metadata.json file containing information specific to that individual. \\

\item Organise data by session (for longitudinal studies) 

If the study is longitudinal (involving multiple data collection points for each participant), create a separate folder within each participant’s directory for each individual participant session. If the study is not longitudinal, skip directly to step 5, placing the task-specific subdirectories (see below). 

The session folders should be named ses-X, where X specifies the session identifier (e.g., ses-01, ses-02, etc.). 

For example, the directory for session 01 of participant 01 would be: 

\begin{lstlisting}[basicstyle=\ttfamily\small,breaklines=true]
/home/username/my_project/participants/part-01/ses-01
\end{lstlisting}

\item Organise data by task/context 

Within the appropriate subdirectory (either the session folder for longitudinal studies or the participant folder for non-longitudinal studies), create subfolders corresponding to the specific context or task of data collection. 

Examples of task/context subdirectory names include interview, fluency, image-description, etc. At least one such subdirectory must exist. The naming of this subdirectory follows no specific guideline and is left to the user. 

A sample directory containing interview data for session 01 of participant 01 in a longitudinal study would therefore look like this: 

\begin{lstlisting}[basicstyle=\ttfamily\small,breaklines=true]
/home/username/my_project/participants/part-01/ses-01/interview
\end{lstlisting}

Or for an equivalent non-longitudinal study: 

\begin{lstlisting}[basicstyle=\ttfamily\small,breaklines=true]
/home/username/my_project/participants/part-01/interview
\end{lstlisting}

\item Store data files 

Place all data files (e.g., audio files, text files of various formats) related to a specific participant, session (if applicable), and task directly within the final task/context subfolder. The naming conventions for such data files are described in the following steps 7-13. 

\end{enumerate}

\subsubsection{Naming conventions}\label{NamingConventions}

LPDS filenames are built from ‘entities’, which are key-value pairs separated by a hyphen (-). Entity keys define what kind of metadata is described, while entity values describe the specific metadata instance.
To follow LPDS filename guidelines separate multiple entities from each other by an underscore (\_). Do not use underscores within an entity’s key or value, as they are reserved for separating distinct entities. 

A comprehensive list of core LPDS entities, their formats, and definitions is provided in Table 1.

\begin{enumerate}[start=7]
\item Include mandatory entities in filenames

Start each filename with the \verb|part| entity (e.g., \verb|part-<label>|) to identify the participant. 

Include the \verb|task| entity (e.g., \verb|task-<label>|) to denote the corresponding task or context of data collection. \\

\item Optional: Include additional entities as needed 

Include optional entities if they are necessary to distinguish between multiple files that would otherwise have the same name or if they add relevant supplementary information. Refer to Table 1 for the format and definition of possible entities. Note: The list is non-exhaustive and other entities may be added if necessary. \\

\item Assemble entity string 

Combine all mandatory and chosen optional entities, separating each key-value pair with an underscore. For the order of the entities in the filename follow the order of entities in Table 1 (entities higher up in the table will come first in the filename). 

Example (entities only): 

\begin{lstlisting}[basicstyle=\ttfamily\small,breaklines=true]
part-01_ses-01_task-interview_cat-motivational_acq-placebo
\end{lstlisting}

\begin{table}[h]
\small
\caption{Core entities of the Language Processing Data Structure (LPDS) are listed in column 1. Each entity consists of a key-value pair as shown in
column 2. Entity descriptions are provided in column 3.}
\label{tab:lpds_entities}
\begin{tabular}{|p{0.1\textwidth}|p{0.2\textwidth}|p{0.63\textwidth}|}
\hline
\textbf{Entity} & \textbf{Format} & \textbf{Description} \\
\hline
\texttt{part} & \texttt{part-<label>} & Participant identifier \\
\hline
\texttt{ses} & \texttt{ses-<label>} & Session identifier \\
\hline
\texttt{task} & \texttt{task-<label>} & Task identifier, denoting experimental task performed \\
\hline
\texttt{cat} & \texttt{cat-<label>} & Category identifier, denoting part-category of performed task \\
\hline
\texttt{acq} & \texttt{acq-<label>} & Acquisition label, specifying variant of same data type \\
\hline
\texttt{run} & \texttt{run-<index>} & Run index, indicating repeated measurements of same acquisition \\
\hline
\texttt{proc} & \texttt{proc-<label>} & Processing label, describing processing applied to the data \\
\hline
\texttt{metric} & \texttt{metric-<label>} & Metric identifier, indicating extracted linguistic metric \\
\hline
\texttt{model} & \texttt{model-<label>} & Model label, identifying model used in data processing \\
\hline
\texttt{group} & \texttt{group-<label>} & Group label, used in group-level analysis \\
\hline
\texttt{param} & \texttt{param-<label>} & Parameter label, denoting specific parameters in processed data \\
\hline
\end{tabular}
\end{table}

\item Optional: Add a suffix

After all entities, include a suffix to further specify the content of the file. Separate the suffix from the entities with an underscore. Examples of common suffixes and their descriptions are provided in Table 2. Note: The list is non-exhaustive and other suffixes may be used to describe the content of the respective file. \\

\begin{table}[h]
\centering
\captionsetup{aboveskip=0pt, belowskip=0pt}
\small
\caption{Example suffixes for naming conventions of the Language Processing Data Structure (LPDS). The suffix is always located immediately before the file extension.}
\label{tab:lpds_suffixes}
\begin{tabular}{|p{0.2\textwidth}|p{0.75\textwidth}|}
\hline
\textbf{Suffix} & \textbf{Description} \\
\hline
\texttt{transcript} & Written transcript of any type of linguistic output \\
\hline
\texttt{text} & Raw or processed text files, typically plain text or pre-tokenised content \\
\hline
\texttt{recording} & Any type of audio recording \\
\hline
\texttt{audio} & Audio files in standard formats (e.g., WAV, MP3), often aligned with transcripts \\
\hline
\texttt{table} & Tabular content such as CSV, TSV, or structured annotation tables \\
\hline
\texttt{embeddings} & Semantic embeddings data, typically vector representations of words or utterances \\
\hline
\texttt{logits} & Model output scores (pre-softmax), often used for classification or uncertainty analysis \\
\hline
\texttt{features} & General-purpose extracted linguistic features (e.g., lexical, syntactic) \\
\hline
\texttt{annotations} & Manual or automatic annotations (e.g., POS tags, discourse markers, errors) \\
\hline
\end{tabular}
\end{table}

\item Add the mandatory file extension 

Conclude the filename with a file extension, which defines the file type (e.g., .txt, .csv, .wav). Precede the file extension with a dot (.). \\

\item Construct the full filename 

Combine the assembled entity string (step 9), the optional suffix (step 10, if used) and the mandatory file extension (step 12).

The general structure for LPDS naming conventions. (Note: Square brackets (\verb|[]|) denote optional parts of the filename):  
\\
\begin{lstlisting}[basicstyle=\ttfamily\small,breaklines=true]
part-<label>[_ses-<label>]_task-<label>[_<key>-<value>][_<suffix>].<extension>
\end{lstlisting}

Example filenames adhering to LPDS naming conventions: 

\begin{lstlisting}[basicstyle=\ttfamily\small,breaklines=true]
part-01_ses-01_task-interview_cat-motivational_acq-placebo_recording.wav

part-12_task-fluency_cat-semantic_acq-animals_transcript.txt
\end{lstlisting}

A sample LPDS structure for a linguistic fluency study is shown in Figure 2. 

\begin{figure}[htbp]
    \centering
    \fcolorbox{black}{white}{\includegraphics[width=1\textwidth]{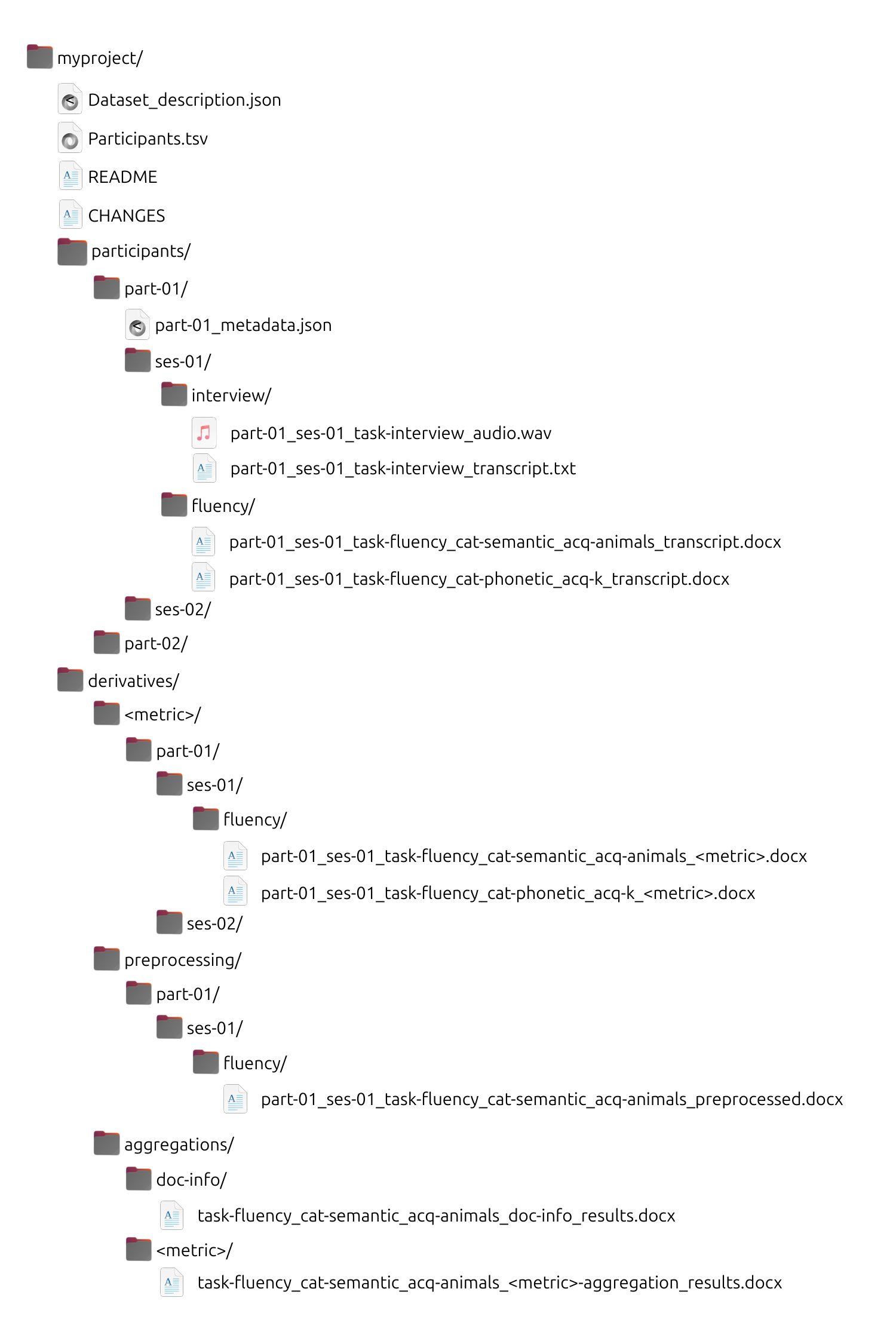}}
    \caption{Example folder structure including introduced naming conventions following the guidelines of the Language Processing Data Structure (LPDS). This example demonstrates storage of language data of a fluency study including participant interviews after linguistic metric extraction.}
    \label{fig:LPDS_example}
\end{figure}

\end{enumerate}

\subsection{Part 2: Pelican\_nlp}\label{pelican}

After organising the data in the LPDS format, follow these steps to install and run the pelican\_nlp package. A general workflow diagram is shown in Figure 3. A list of all possible parameters, a description and a usage guide are provided in Table S1 in the supplementary information. This table should be consulted when adapting a configuration file. Note: This guide is optimised for Linux-based operating systems (e.g., Ubuntu). Details might differ for other operating systems like Windows or MacOS. 

\begin{enumerate}

\item Prepare configuration file 

Prepare a single .yml configuration file to define the entire processing pipeline. Either choose a pre-defined configuration file provided in the projects GitHub repository (\url{https://github.com/ypauli/pelican\_nlp}) (option A), replicate a study by using a specifically to the study adapted configuration file (option B), or create a custom configuration file from scratch (option C). \\
\\

\textbf{Option A: Choose pre-defined configuration file} \\

\begin{enumerate}[label=\roman*.]
\item Choose a sample configuration file 

We provide sample configuration files for the sample use-cases described in section \ref{Development} (uncertainty modelling in multimodal speech\cite{rohanian2025uncertaintymodelingmultimodalspeech}, a semantic fluency analysis, the analysis of artificially generated language and speech-based prediction in psychosis). These sample configuration files can be found in the sample\_configuration\_files folder of the pelican\_nlp GitHub repository (\url{https://github.com/ypauli/pelican\_nlp}). \\

\item Adapt the configuration 

Change necessary parameters to adapt the configuration file. Consult Table S1 in supplementary information for adaptation options. \\

\end{enumerate} 

\textbf{Option B: Use a specifically adapted configuration file} \\

\begin{enumerate}[label=\roman*.]
\item Choose a configuration file set up by a colleague 

If the goal is to precisely replicate a colleagues's results or those from a previous publication, obtain the exact config.yml file used to define the processing pipeline. \\

\item Adapt the configuration 

Change necessary parameters to adapt the configuration file to the data to be used. Consult Table S1 in the supplementary information for adaptation options. However, a configuration file that has specifically been adapted for a specific study is normally chosen to replicate the results, and therefore changes that alter the core processing pipeline should be made with caution to ensure replication validity. \\

\end{enumerate} 

\textbf{Option C: Create custom configuration file from scratch} \\

\begin{enumerate}[label=\roman*.]
\item Download the general configuration file 

Retrieve the general\_configurations.yml file from the pelican\_nlp GitHub repository (\url{https://github.com/ypauli/pelican\_nlp}). This file contains all possible parameters the pelican\_nlp package can utilise. \\

\item Adapt the configuration 

Modify the parameters based on the pipeline requirements. Consult Table S1 in supplementary information for adaptation options. Remove unused parameters, provided they are not marked as required within the supplementary Table S1. \\

\end{enumerate}

\item Place the configuration file 

Save the chosen and adapted configuration file from step 1 to your main project directory (directory according to step 1 of Part 1: LPDS). \\

\item Navigate to the main project directory in the command line 

To open a terminal in the project directory, right-click the project directory and select "Open in Terminal". All following commands will be executed from the command line of this terminal. \\

\item Set up project environment

To avoid conflicts with other packages, it is strongly recommended to create a dedicated Conda environment. In case Conda is not installed on your device execute the following commands to install Conda. If Conda is already installed skip this step.

\begin{lstlisting}[basicstyle=\ttfamily\small,breaklines=true]
wget https://repo.anaconda.com/archive/Anaconda3-2022.05-Linux-x86_64.sh
bash Anaconda3-2022.05-Linux-x86_64.sh
\end{lstlisting}

When Conda is installed on your device execute the following two commands in the command line to create and activate a Conda environment named pelican-nlp with Python version 3.10:

\begin{lstlisting}[basicstyle=\ttfamily\small,breaklines=true]
conda create --name pelican-nlp --channel defaults python=3.10
conda activate pelican-nlp
\end{lstlisting}

\item Install pelican\_nlp package

Install the pelican\_nlp package to the activated Conda environment by executing the following commands in your command line:

\begin{lstlisting}[basicstyle=\ttfamily\small,breaklines=true]
python -m pip --version
conda install pip
pip install pelican-nlp
\end{lstlisting}

\item Execute the pelican\_nlp pipeline

Run the pelican\_nlp pipeline with the chosen configuration file by executing the following command in your command line:

\begin{lstlisting}[basicstyle=\ttfamily\small,breaklines=true]
pelican-run
\end{lstlisting}

This command will automatically find the specified configuration file and your data folder, execute all specified steps, and create a new folder called derivatives in your main project directory containing all the computed results. This is shown in Figure \ref{fig:LPDS_example}.

\end{enumerate}

\begin{figure}[htbp]
    \centering
    \fcolorbox{black}{white}{\includegraphics[width=1\textwidth]{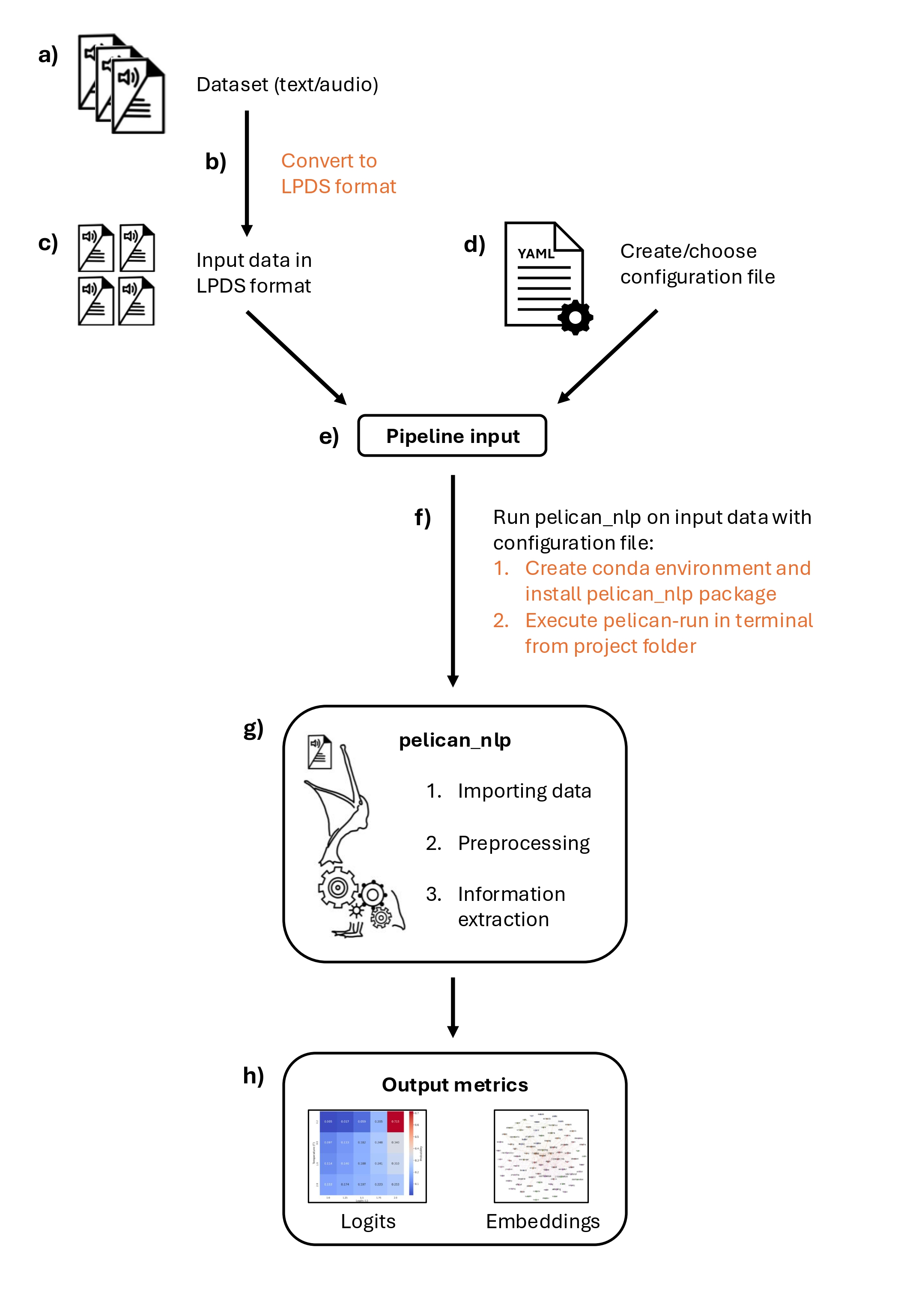}}
    \caption{Workflow diagram illustrating how to use the pelican\_nlp package. Orange boxes highlight the steps in which user input is required. The original dataset (a) is transformed into Language Processing Data Structure (LPDS) format (b). The transformed dataset (c) and the created/chosen configuration file (d) represent the pipeline input (e). Executing the terminal command pelican-run package on the pipeline input (f) executes the pelican\_nlp package (g). The pelican\_nlp package will then calculate and output linguistic metrics (h) and store them in csv format.}
    \label{fig:workflow}
\end{figure}

\section{Timing}\label{Timing}

Below are estimated time requirements for each stage of the protocol. Durations may vary depending on the system configurations, dataset size, and selected processing options. \\

\textbf{Part 1: LPDS structure setup — \(\sim \) 1 h} \\
\\
Step 1, establish main project directory — \(\sim \) 5 min \\
\\
Step 2, create core project subdirectories — \(\sim \) 5 min \\
\\
Steps 3–5, organise data — \(\sim \) 20 min \\
\\
Step 6, store data files — \(\sim \) 15 min \\
\\
Steps 7–13, apply LPDS naming conventions — \(\sim \) 15 min \\
\\

\textbf{Part 2: Pelican NLP setup and execution — 1–4 h (depending on data volume, processing pipeline and hardware setup)} \\
\\
Step 1, prepare configuration file: \\
Option A or B — 10–30 min; \\
Option C (custom config) — 1–2 h \\
\\
Step 2, place configuration file in project directory — \(\sim \) 1 min \\
\\
Step 3, navigate to project directory in terminal — \(\sim \) 1 min \\
\\
Step 4, set up Conda environment — \(\sim \) 5 min \\
\\
Step 5, install pelican-nlp package — \(\sim \) 5 min \\
\\
Step 6, execute pipeline by running \verb|pelican-run|: 1-15 min per file \\
text-only cleaning — \texttt{<} 1 min per file; \\
embedding extraction — \(\sim \) 2 min per file; \\
logits extraction - \(\sim \) 5 min per file; \\
audio feature extraction - \(\sim \) 1 min per recording minute;

\section{Troubleshooting}\label{sec5}

Occasional system crashes or freezes may occur while running pelican\_nlp, particularly during computationally intensive tasks. These issues are caused by GPU memory limitations. To resolve this, users are advised to run pelican\_nlp on a system with sufficient available VRAM (ideally at least 16 GB) for stable performance when using larger models or complex configurations.

\section{Anticipated results}\label{AnticipatedResults}

Upon successful execution of the pelican\_nlp package via the pelican-run command within the main project directory, a new subdirectory named ‘derivatives’ will be created (if one does not already exist). This derivatives directory serves as the central repository for all outputs generated by the specified processing pipeline. The exact content and structure within the derivatives folder are directly determined by the parameters set in the config.yml file, ensuring that the outputs precisely reflect the chosen analytical protocol. For an example derivatives folder see Fig. \ref{fig:LPDS_example}.

\subsection{Key characteristics of the output}\label{KeyCharacteristicsOutput}

\begin{itemize}
    \item LPDS compliance: Crucially, the organisation of files and subdirectories within the derivatives folder adheres to LPDS naming conventions and structural principles. This means that output filenames will include relevant entities (e.g., \verb|part-|, \verb|ses-|, \verb|task-|, \verb|proc-|, \verb|metric-|, \verb|model-|) mirroring the input data, facilitating clear linkage between raw data, processed data, and extracted features. 
    \item Modularity: Outputs are typically organised into distinct subdirectories based on the type of processing or feature extraction performed. 
\end{itemize}

\subsection{Common subdirectories and their contents}\label{SubdirAndContent}

The derivatives directory may contain any of the following subdirectories, depending on the user-specified settings in the config.yml file: 

\begin{enumerate}
\item preprocessing/:

This subdirectory will contain data files that have undergone one or more preprocessing steps but are not yet final feature extractions. 

Examples: Cleaned text files (e.g., \verb|[...]_proc-cleaned_text.txt|) with timestamps removed, punctuation standardised, or speaker tags harmonised. 

The specific \verb|proc-<label>| in the filename will indicate the type of preprocessing applied. \\

\item Metric-Specific Subdirectories (e.g., \verb|embeddings/|, \verb|similarity/|, \verb|logits/|, \verb|distance_from_randomness/|): 

For each linguistic metric extracted, pelican\_nlp will create a dedicated subdirectory named after that metric. 

These folders will contain the actual feature data, typically saved in common machine-readable formats. 

Examples: 

\begin{lstlisting}[basicstyle=\ttfamily\small,breaklines=true]
embeddings/part-01_task-interview_metric-embeddings_model-bert_description-mean.csv
\end{lstlisting}

This example file contains would word or sentence embeddings.

\begin{lstlisting}[basicstyle=\ttfamily\small,breaklines=true]
similarity/part-01_task-fluency_metric-similarity_model-bert_description-window.json
\end{lstlisting}

This example file would contain semantic similarity scores. 

\begin{lstlisting}[basicstyle=\ttfamily\small,breaklines=true]
logits/part-01_task-interview_metric-logits_model-RoBERTa.pt
\end{lstlisting}

This example file would contain raw model logits. 

The adaptable filenames will clearly indicate the source data and the parameters of the extraction (e.g., participant, task, model used, any aggregation method or specific descriptor for the metric). \\

\item aggregations/: 

This subdirectory will contain files that aggregate results across multiple participants, sessions, or files, often for user convenience or as input for group-level statistical analyses. 

Example: 
A single CSV file compiling a specific derived metric for all participants in a study:

\begin{lstlisting}[basicstyle=\ttfamily\small,breaklines=true]
acq-animals_metric-similarity_results-aggregated_table.csv
\end{lstlisting}

The nature and format of these aggregations are defined by the configuration. \\

\item Logs and Reports:

Depending on the configuration, pelican\_nlp may also output processing logs or summary reports in a subfolder ‘logs’ within the derivatives directory. These can be valuable for troubleshooting and documenting the execution details. 

Examples for logs and summary files:

\begin{lstlisting}[basicstyle=\ttfamily\small,breaklines=true]
pelican_log.txt
processing_summary.html
\end{lstlisting}

\end{enumerate}

\subsection{Facilitating downstream analysis and reproducibility}\label{FacilitatingDownstreamAnalysisAndReproducibility} 

The standardised, LPDS-compliant structure of the derivatives directory, coupled with descriptive filenames, ensures that the outputs are not only interpretable but also readily usable for subsequent statistical analysis, machine learning model training, visualisation, and, critically, for replication by other researchers using the same pelican\_nlp version and configuration file.

Sample derivatives directory structures and example output files for common use cases (e.g., fluency analysis, discourse analysis) are provided in the project's GitHub repository (\url{https://github.com/ypauli/pelican\_nlp}) to further illustrate the anticipated results.


\bibliography{sn-bibliography}

@article{crema2022nlp_clinical,
 author  = "Crema, C. and Attardi, G. and Sartiano, D. and Redolfi, A.",
 title   = "Natural language processing in clinical neuroscience and psychiatry: A review",
 journal  = "Front. Psychiatry",
 volume  = "13",
 year   = "2022"
}

@article{feuerriegel2025nlp_behavioral,
 author  = "Feuerriegel, S. and Maarouf, A. and Bär, D. and others",
 title   = "Using natural language processing to analyse text data in behavioural science",
 journal  = "Nat Rev Psychol",
 volume  = "4",
 pages   = "96--111",
 year   = "2025"
}

@inproceedings{vaswani2017attention,
  title     = {Attention Is All You Need},
  author    = {Vaswani, Ashish and Shazeer, Noam and Parmar, Niki and Uszkoreit, Jakob and Jones, Llion and Gomez, Aidan N. and Kaiser, {\L}ukasz and Polosukhin, Illia},
  booktitle = {Proceedings of the 31st International Conference on Neural Information Processing Systems},
  pages     = {5998--6008},
  year      = {2017},
  publisher = {Curran Associates, Inc.},
  address   = {Red Hook, NY, USA},
  url       = {https://arxiv.org/abs/1706.03762}
}

@book{bird2009nlp_python,
 author  = "Bird, S. and Klein, E. and Loper, E.",
 title   = "Natural Language Processing with Python",
 publisher = "O’Reilly Media",
 address   = "Sebastopol, CA, USA", 
 year   = "2009"
}

@inproceedings{manning2014corenlp,
 author  = "Manning, C. D. and others",
 title   = "The Stanford CoreNLP Natural Language Processing Toolkit",
 booktitle = "Proceedings of the 52nd Annual Meeting of the Association for Computational Linguistics (ACL): System Demonstrations",
 pages   = "55--60",
 year   = "2014"
}

@misc{honnibal2020spacy,
 author  = "Honnibal, M. and Montani, I. and Van Landeghem, S. and Boyd, A.",
 title   = "spaCy: Industrial‑strength Natural Language Processing in Python",
 howpublished = "Zenodo",
 year   = "2020",
 doi    = "10.5281/zenodo.1212303"
}

@inproceedings{akbik2019flair,
 author  = "Akbik, A. and Bergmann, T. and Blythe, D. and Rasul, K. and Schweter, S. and Vollgraf, R.",
 title   = "FLAIR: An Easy‑to‑Use Framework for State‑of‑the‑Art NLP",
 booktitle = "Proceedings of the 2019 Conference of the North American Chapter of the Association for Computational Linguistics (Demonstrations)",
 pages   = "54--59",
 year   = "2019",
 address  = "Minneapolis, Minnesota",
 publisher = "Association for Computational Linguistics"
}

@misc{fokkens2013offspring,
 author  = "Fokkens, A. and van Erp, M. and Postma, M. and Pedersen, T. and Vossen, P. and Freire, N.",
 title   = "Offspring from reproduction problems: What replication failure teaches us",
 booktitle = "Proceedings of the 51st Annual Meeting of the Association for Computational Linguistics",
 pages   = "1691--1701",
 year   = "2013",
 publisher = "Association for Computational Linguistics"
}

@article{munafo2017manifesto,
 author  = "Munafò, M. R. and others",
 title   = "A manifesto for reproducible science",
 journal  = "Nat Hum Behav",
 volume  = "1",
 pages   = "0021",
 year   = "2017"
}

@article{baker2016reproducibility_survey,
 author  = "Baker, M.",
 title   = "1,500 scientists lift the lid on reproducibility",
 journal  = "Nature",
 volume  = "533",
 pages   = "452--454",
 year   = "2016"
}

@article{ioannidis2005why_most_false,
 author  = "Ioannidis, J. P. A.",
 title   = "Why Most Published Research Findings Are False",
 journal  = "PLoS Med",
 volume  = "2",
 pages   = "e124",
 year   = "2005"
}

@article{denny2018text_preprocessing,
 author  = "Denny, M. J. and Spirling, A.",
 title   = "Text preprocessing for unsupervised learning: Why it matters, when it misleads, and what to do about it",
 journal  = "Political Analysis",
 volume  = "26",
 number  = "2",
 pages   = "168--189",
 year   = "2018",
 doi    = "10.1017/pan.2017.44"
}

@article{botvinik2020variability,
 author  = "Botvinik‑Nezer, R. and Holzmeister, F. and Camerer, C. F. and others",
 title   = "Variability in the analysis of a single neuroimaging dataset by many teams",
 journal  = "Nature",
 volume  = "582",
 pages   = "84--88",
 year   = "2020"
}

@article{silberzahn2018many_analysts,
 author  = "Silberzahn, R. and Uhlmann, E. L. and Martin, D. P. and others",
 title   = "Many Analysts, One Data Set: Making Transparent How Variations in Analytic Choices Affect Results",
 journal  = "Advances in Methods and Practices in Psychological Science",
 volume  = "1",
 number  = "3",
 pages   = "337--356",
 year   = "2018",
 doi    = "10.1177/2515245917747646"
}

@article{carp2012analytic_flexibility,
 author  = "Carp, J.",
 title   = "On the Plurality of (Methodological) Worlds: Estimating the Analytic Flexibility of fMRI Experiments",
 journal  = "Front. Neurosci.",
 volume  = "6",
 year   = "2012"
}

@article{wicherts2016degrees_freedom,
 author  = "Wicherts, J. M. and others",
 title   = "Degrees of Freedom in Planning, Running, Analyzing, and Reporting Psychological Studies: A Checklist to Avoid p‑Hacking",
 journal  = "Front. Psychol.",
 volume  = "7",
 year   = "2016"
}

@article{simmons2011false_positive_psychology,
 author  = "Simmons, J. P. and Nelson, L. D. and Simonsohn, U.",
 title   = "False‑Positive Psychology: Undisclosed Flexibility in Data Collection and Analysis Allows Presenting Anything as Significant",
 journal  = "Psychological Science",
 volume  = "22",
 number  = "11",
 pages   = "1359--1366",
 year   = "2011",
 doi    = "10.1177/0956797611417632"
}

@article{schweinsberg2021same_data,
 author  = "Schweinsberg, M. and others",
 title   = "Same data, different conclusions: Radical dispersion in empirical results when independent analysts operationalize and test the same hypothesis",
 journal  = "Organizational Behavior and Human Decision Processes",
 volume  = "165",
 pages   = "228--249",
 year   = "2021"
}

@article{gorgolewski2016bids,
 author  = "Gorgolewski, K. and Auer, T. and Calhoun, V. and others",
 title   = "The brain imaging data structure, a format for organizing and describing outputs of neuroimaging experiments",
 journal  = "Sci Data",
 volume  = "3",
 pages   = "160044",
 year   = "2016"
}

@article{poldrack2017repro_neuroimaging,
 author  = "Poldrack, R. A. and others",
 title   = "Scanning the horizon: towards transparent and reproducible neuroimaging research",
 journal  = "Nat Rev Neurosci",
 volume  = "18",
 pages   = "115--126",
 year   = "2017"
}

@article{esteban2019fmri_prep,
 author  = "Esteban, O. and Markiewicz, C. J. and Blair, R. W. and others",
 title   = "fMRIPrep: a robust preprocessing pipeline for functional MRI",
 journal  = "Nat Methods",
 volume  = "16",
 pages   = "111--116",
 year   = "2019"
}

@article{peng2011reproducible_computational_science,
 author  = "Peng, R. D.",
 title   = "Reproducible Research in Computational Science",
 journal  = "Science",
 volume  = "334",
 pages   = "1226--1227",
 year   = "2011"
}

@article{sandve2013ten_rules,
 author  = "Sandve, G. K. and others",
 title   = "Ten simple rules for reproducible computational research",
 journal  = "PLoS Comput Biol",
 volume  = "9",
 pages   = "e1003285",
 year   = "2013"
}

@article{barnes2010publish_code,
 author  = "Barnes, N.",
 title   = "Publish your computer code: it is good enough",
 journal  = "Nature",
 volume  = "467",
 pages   = "753",
 year   = "2010",
 doi    = "10.1038/467753a"
}

@article{bojanowski2017subword,
 author  = "Bojanowski, P. and Grave, E. and Joulin, A. and Mikolov, T.",
 title   = "Enriching Word Vectors with Subword Information",
 journal  = "TACL",
 volume  = "5",
 pages   = "135--146",
 year   = "2017"
}

@inproceedings{qi2020stanza,
 author  = "Qi, P. and Zhang, Y. and Zhang, Y. and Bolton, J. and Manning, C. D.",
 title   = "Stanza: A Python Natural Language Processing Toolkit for Many Human Languages",
 booktitle = "Association for Computational Linguistics (ACL) System Demonstrations",
 year   = "2020"
}

@inproceedings{wolf2020transformers,
 author  = "Wolf, T. and others",
 title   = "Transformers: State‑of‑the‑Art Natural Language Processing",
 booktitle = "Proceedings of the 2020 Conference on Empirical Methods in Natural Language Processing (EMNLP): System Demonstrations",
 pages   = "38--45",
 year   = "2020"
}

@inproceedings{devlin2019bert,
 author  = "Devlin, J. and Chang, M.-W. and Lee, K. and Toutanova, K.",
 title   = "BERT: Pre‑training of Deep Bidirectional Transformers for Language Understanding",
 booktitle = "Proceedings of the 2019 Conference of the North American Chapter of the Association for Computational Linguistics (NAACL‑HLT)",
 pages   = "4171--4186",
 year   = "2019"
}

@misc{liu2019roberta,
 author  = "Liu, Y. and others",
 title   = "RoBERTa: A Robustly Optimized BERT Pretraining Approach",
 year   = "2019",
 url = {https://arxiv.org/abs/1907.11692}
}

@misc{touvron2023llama,
 author  = "Touvron, H. and others",
 title   = "LLaMA: Open and Efficient Foundation Language Models",
 year   = "2023",
 url = {https://arxiv.org/abs/2302.13971}
}

@inproceedings{gardner2018allennlp,
 author  = "Gardner, M. and Grus, J. and Neumann, M. and Tafjord, O. and Dasigi, P. and Liu, N. F. and Peters, M. and Schmitz, M. and Zettlemoyer, L.",
 title   = "AllenNLP: A Deep Semantic Natural Language Processing Platform",
 booktitle = "Proceedings of Workshop for NLP Open Source Software (NLP‑OSS)",
 pages   = "1--6",
 year   = "2018",
 address  = "Melbourne, Australia",
 publisher = "Association for Computational Linguistics"
}

@article{wieling2018repro_computational_linguistics,
 author  = "Wieling, M. and Rawee, J. and van Noord, G.",
 title   = "Reproducibility in Computational Linguistics: Are We Willing to Share?",
 journal  = "Computational Linguistics",
 volume  = "44",
 number  = "4",
 pages   = "641--649",
 year   = "2018",
 doi    = "10.1162/coli_a_00330"
}

@article{schofield2016stemmers_topic_models,
 author  = "Schofield, A. and Mimno, D.",
 title   = "Comparing Apples to Apple: The Effects of Stemmers on Topic Models",
 journal  = "Transactions of the Association for Computational Linguistics",
 volume  = "4",
 pages   = "287--300",
 year   = "2016",
 doi    = "10.1162/tacl_a_00099"
}

@article{crane2018questionable_qa,
 author  = "Crane, M.",
 title   = "Questionable Answers in Question Answering Research: Reproducibility and Variability of Published Results",
 journal  = "Transactions of the Association for Computational Linguistics",
 volume  = "6",
 pages   = "241--252",
 year   = "2018",
 doi    = "10.1162/tacl_a_00018"
}

@inproceedings{mieskes2017quantitative_acl_ethics,
 author  = "Mieskes, M.",
 title   = "A Quantitative Study of Data in the NLP community",
 booktitle = "Proceedings of the First ACL Workshop on Ethics in Natural Language Processing",
 pages   = "23--29",
 year   = "2017",
 address  = "Valencia, Spain",
 publisher = "Association for Computational Linguistics"
}

@article{belz2022metrological_repro_lab,
 author  = "Belz, A.",
 title   = "A Metrological Perspective on Reproducibility in NLP*",
 journal  = "Computational Linguistics",
 volume  = "48",
 number  = "4",
 pages   = "1125--1135",
 year   = "2022",
 doi    = "10.1162/coli_a_00448"
}

@inproceedings{mieskes2019community_replication_ranlp,
 author  = "Mieskes, M. and Fort, K. and Névéol, A. and Grouin, C. and Cohen, K.",
 title   = "Community Perspective on Replicability in Natural Language Processing",
 booktitle = "Proceedings of the International Conference on Recent Advances in Natural Language Processing (RANLP 2019)",
 pages   = "768--775",
 year   = "2019",
 address  = "Varna, Bulgaria",
 publisher = "INCOMA Ltd."
}

@inproceedings{belz2021reprogen_shared_task_2021,
 author  = "Belz, A. and Shimorina, A. and Agarwal, S. and Reiter, E.",
 title   = "The ReproGen Shared Task on Reproducibility of Human Evaluations in NLG: Overview and Results",
 booktitle = "Proceedings of the 14th International Conference on Natural Language Generation",
 pages   = "249--258",
 year   = "2021",
 address  = "Aberdeen, Scotland, UK",
 publisher = "Association for Computational Linguistics"
}

@inproceedings{branco2020reprolang2020,
 author  = "Branco, A. and Calzolari, N. and Vossen, P. and Van Noord, G. and van Uytvanck, D. and Silva, J. and Gomes, L. and Moreira, A. and Elbers, W.",
 title   = "A Shared Task of a New, Collaborative Type to Foster Reproducibility: A First Exercise in the Area of Language Science and Technology with REPROLANG2020",
 booktitle = "Proceedings of the Twelfth Language Resources and Evaluation Conference",
 pages   = "5539--5545",
 year   = "2020",
 address  = "Marseille, France",
 publisher = "European Language Resources Association"
}

@inproceedings{belz2024repronlp_shared_task_2024,
 author  = "Belz, A. and Thomson, C.",
 title   = "The 2024 ReproNLP Shared Task on Reproducibility of Evaluations in NLP: Overview and Results",
 booktitle = "Proceedings of the Fourth Workshop on Human Evaluation of NLP Systems (HumEval) @ LREC‑COLING 2024",
 pages   = "91--105",
 year   = "2024",
 address  = "Torino, Italia",
 publisher = "ELRA and ICCL"
}

@inproceedings{belz2021systematic_review_reproducibility_nlp,
 author  = "Belz, A. and Shimorina, A. and Agarwal, S. and Reiter, E.",
 title   = "A Systematic Review of Reproducibility Research in Natural Language Processing",
 booktitle = "Proceedings of the 16th Conference of the European Chapter of the Association for Computational Linguistics",
 pages   = "381--393",
 year   = "2021",
 address  = "Online",
 publisher = "Association for Computational Linguistics"
}

@inproceedings{howcroft2020twenty_years_confusion_nlg,
 author  = "Howcroft, D. M. and Belz, A. and Clinciu, M.-A. and Gkatzia, D. and Hasan, S. A. and Mahamood, S. and Mille, S. and van Miltenburg, E. and Santhanam, S. and Rieser, V.",
 title   = "Twenty Years of Confusion in Human Evaluation: NLG Needs Evaluation Sheets and Standardised Definitions",
 booktitle = "Proceedings of the 13th International Conference on Natural Language Generation",
 pages   = "169--182",
 year   = "2020",
 address  = "Dublin, Ireland",
 publisher = "Association for Computational Linguistics"
}

@inproceedings{belz2022repronlp_reprogen2022,
 author  = "Belz, A. and Shimorina, A. and Popovic, M. and Reiter, E.",
 title   = "The 2022 reprogen shared task on reproducibility of evaluations in nlg: Overview and results",
 booktitle = "INLG 2022",
 pages   = "43",
 year   = "2022"
}

@inproceedings{belz2023repronlp_shared_task_2023,
 author  = "Belz, A. and Thomson, C.",
 title   = "The 2023 ReproNLP shared task on reproducibility of evaluations in NLP: Overview and results",
 booktitle = "Proceedings of the 3rd Workshop on Human Evaluation of NLP Systems",
 pages   = "35--48",
 year   = "2023",
 address  = "Varna, Bulgaria",
 publisher = "INCOMA Ltd."
}

@inproceedings{belz2023missing_info_negative_results_nlp,
 author  = "Belz, A. and Thomson, C. and Reiter, E. and Abercrombie, G. and Alonso‑Moral, J. M. and Arvan, M. and Braggaar, A. and Cieliebak, M. and Clark, E. and van Deemter, K. and Dinkar, T. and Dušek, O. and Eger, S. and Fang, Q. and Gao, M. and Gatt, A. and Gkatzia, D. and González‑Corbelle, J. and Hovy, D. and Hürlimann, M. and Ito, T. and Kelleher, J. D. and Klubicka, F. and Krahmer, E. and Lai, H. and van der Lee, C. and Li, Y. and Mahamood, S. and Mieskes, M. and van Miltenburg, E. and Mosteiro, P. and Nissim, M. and Parde, N. and Plátek, O. and Rieser, V. and Ruan, J. and Tetreault, J. and Toral, A. and Wan, X. and Wanner, L. and Watson, L. and Yang, D.",
 title   = "Missing information, unresponsive authors, experimental flaws: The impossibility of assessing the reproducibility of previous human evaluations in NLP",
 booktitle = "Proceedings of the Fourth Workshop on Insights from Negative Results in NLP",
 pages   = "1--10",
 year   = "2023",
 address  = "Dubrovnik, Croatia"
}

@misc{falcon2019pytorch,
  author       = {William Falcon and PyTorch Lightning Team},
  title        = {PyTorch Lightning},
  year         = {2019},
  url = {https://github.com/Lightning-AI/pytorch-lightning}
}

@misc{mikolov2013efficientestimationwordrepresentations,
   title={Efficient Estimation of Word Representations in Vector Space}, 
   author={Tomas Mikolov and Kai Chen and Greg Corrado and Jeffrey Dean},
   year={2013},
   eprint={1301.3781},
   archivePrefix={arXiv},
   primaryClass={cs.CL},
   url={https://arxiv.org/abs/1301.3781}, 
}

@article{
doi:10.1073/pnas.2203150119,
author = {Nate Breznau and Eike Mark Rinke and Alexander Wuttke and Hung H. V. Nguyen and Muna Adem and Jule Adriaans and Amalia Alvarez-Benjumea and Henrik K. Andersen and Daniel Auer and Flavio Azevedo and Oke Bahnsen and Dave Balzer and Gerrit Bauer and Paul C. Bauer and Markus Baumann and Sharon Baute and Verena Benoit and Julian Bernauer and Carl Berning and Anna Berthold and Felix S. Bethke and Thomas Biegert and Katharina Blinzler and Johannes N. Blumenberg and Licia Bobzien and Andrea Bohman and Thijs Bol and Amie Bostic and Zuzanna Brzozowska and Katharina Burgdorf and Kaspar Burger and Kathrin B. Busch and Juan Carlos-Castillo and Nathan Chan and Pablo Christmann and Roxanne Connelly and Christian S. Czymara and Elena Damian and Alejandro Ecker and Achim Edelmann and Maureen A. Eger and Simon Ellerbrock and Anna Forke and Andrea Forster and Chris Gaasendam and Konstantin Gavras and Vernon Gayle and Theresa Gessler and Timo Gnambs and Amélie Godefroidt and Max Grömping and Martin Groß and Stefan Gruber and Tobias Gummer and Andreas Hadjar and Jan Paul Heisig and Sebastian Hellmeier and Stefanie Heyne and Magdalena Hirsch and Mikael Hjerm and Oshrat Hochman and Andreas Hövermann and Sophia Hunger and Christian Hunkler and Nora Huth and Zsófia S. Ignácz and Laura Jacobs and Jannes Jacobsen and Bastian Jaeger and Sebastian Jungkunz and Nils Jungmann and Mathias Kauff and Manuel Kleinert and Julia Klinger and Jan-Philipp Kolb and Marta Kołczyńska and John Kuk and Katharina Kunißen and Dafina Kurti Sinatra and Alexander Langenkamp and Philipp M. Lersch and Lea-Maria Löbel and Philipp Lutscher and Matthias Mader and Joan E. Madia and Natalia Malancu and Luis Maldonado and Helge Marahrens and Nicole Martin and Paul Martinez and Jochen Mayerl and Oscar J. Mayorga and Patricia McManus and Kyle McWagner and Cecil Meeusen and Daniel Meierrieks and Jonathan Mellon and Friedolin Merhout and Samuel Merk and Daniel Meyer and Leticia Micheli and Jonathan Mijs and Cristóbal Moya and Marcel Neunhoeffer and Daniel Nüst and Olav Nygård and Fabian Ochsenfeld and Gunnar Otte and Anna O. Pechenkina and Christopher Prosser and Louis Raes and Kevin Ralston and Miguel R. Ramos and Arne Roets and Jonathan Rogers and Guido Ropers and Robin Samuel and Gregor Sand and Ariela Schachter and Merlin Schaeffer and David Schieferdecker and Elmar Schlueter and Regine Schmidt and Katja M. Schmidt and Alexander Schmidt-Catran and Claudia Schmiedeberg and Jürgen Schneider and Martijn Schoonvelde and Julia Schulte-Cloos and Sandy Schumann and Reinhard Schunck and Jürgen Schupp and Julian Seuring and Henning Silber and Willem Sleegers and Nico Sonntag and Alexander Staudt and Nadia Steiber and Nils Steiner and Sebastian Sternberg and Dieter Stiers and Dragana Stojmenovska and Nora Storz and Erich Striessnig and Anne-Kathrin Stroppe and Janna Teltemann and Andrey Tibajev and Brian Tung and Giacomo Vagni and Jasper Van Assche and Meta van der Linden and Jolanda van der Noll and Arno Van Hootegem and Stefan Vogtenhuber and Bogdan Voicu and Fieke Wagemans and Nadja Wehl and Hannah Werner and Brenton M. Wiernik and Fabian Winter and Christof Wolf and Yuki Yamada and Nan Zhang and Conrad Ziller and Stefan Zins and Tomasz Żółtak},
title = {Observing many researchers using the same data and hypothesis reveals a hidden universe of uncertainty},
journal = {Proceedings of the National Academy of Sciences},
volume = {119},
number = {44},
pages = {e2203150119},
year = {2022},
doi = {10.1073/pnas.2203150119},
URL = {https://www.pnas.org/doi/abs/10.1073/pnas.2203150119},
eprint = {https://www.pnas.org/doi/pdf/10.1073/pnas.2203150119}}

@misc{rohanian2025uncertaintymodelingmultimodalspeech,
   title={Uncertainty Modeling in Multimodal Speech Analysis Across the Psychosis Spectrum}, 
   author={Morteza Rohanian and Roya M. Hüppi and Farhad Nooralahzadeh and Noemi Dannecker and Yves Pauli and Werner Surbeck and Iris Sommer and Wolfram Hinzen and Nicolas Langer and Michael Krauthammer and Philipp Homan},
   year={2025},
   eprint={2502.18285},
   archivePrefix={arXiv},
   primaryClass={cs.CL},
   url={https://arxiv.org/abs/2502.18285}, 
}

@article{article,
author = {Forkel, Robert and List, Johann-Mattis and Greenhill, Simon and Rzymski, Christoph and Bank, Sebastian and Cysouw, Michael and Hammarström, Harald and Haspelmath, Martin and Kaiping, Gereon and Gray, Russell},
year = {2018},
month = {10},
pages = {180205},
title = {Cross-Linguistic Data Formats, advancing data sharing and re-use in comparative linguistics},
volume = {5},
journal = {Scientific Data},
doi = {10.1038/sdata.2018.205}
}

@article{wilkinson2016fair,
 author    = {Mark D. Wilkinson and Michel Dumontier and IJsbrand Jan Aalbersberg and Gabrielle Appleton and Myles Axton and Arie Baak and Niklas Blomberg and Jan-Willem Boiten and Luiz Bonino da Silva Santos and Philip E. Bourne and Jim Brookes and Tim Clark and Mercè Crosas and Ingrid Dillo and Olivier Dumon and Scott Edmunds and Chris T. Evelo and Richard Finkers and Alejandra Gonzalez-Beltran and Alasdair J.G. Gray and Paul Groth and Carole Goble and Jeffrey S. Grethe and Jaap Heringa and Peter A.C. ‘t Hoen and Rob Hooft and Tobias Kuhn and Ruben Kok and Joost Kok and Scott J. Lusher and Maryann E. Martone and Albert Mons and Abel L. Packer and Bengt Persson and Philippe Rocca-Serra and Marco Roos and Rene van Schaik and Susanna-Assunta Sansone and Erik Schultes and Thierry Sengstag and Ted Slater and George Strawn and Morris A. Swertz and Mark Thompson and Johan van der Lei and Erik van Mulligen and Jan Velterop and Peter Wittenburg and Katherine Wolstencroft and Jun Zhao and Barend Mons},
 title    = {The {FAIR} Guiding Principles for scientific data management and stewardship},
 journal   = {Scientific Data},
 volume    = {3},
 pages    = {160018},
 year     = {2016},
 doi     = {10.1038/sdata.2016.18},
 url     = {https://doi.org/10.1038/sdata.2016.18}
}

@inproceedings{Eyben2010openSMILE,
 author  = {Florian Eyben and Martin W{\"o}llmer and Bj{\"o}rn Schuller},
 title   = {openSMILE: The Munich Versatile and Fast Open-Source Audio Feature Extractor},
 booktitle = {Proceedings of the 18th ACM International Conference on Multimedia (MM)},
 year   = {2010},
 pages   = {1459--1462},
 publisher = {ACM},
 address  = {Florence, Italy},
 isbn   = {978-1-60558-933-6}
}

@inproceedings{Mertens2004Prosogram,
 author  = {Mertens, Piet},
 title   = {The Prosogram: Semi-Automatic Transcription of Prosody Based on a Tonal Perception Model},
 booktitle = {Proceedings of Speech Prosody 2004},
 year   = {2004},
 pages   = {549--552},
 address  = {Nara, Japan}
}

@inproceedings{Joulin2016fastText,
 author  = {Joulin, Armand and Grave, Edouard and Bojanowski, Piotr and Mikolov, Tomas},
 title   = {Bag of Tricks for Efficient Text Classification},
 booktitle = {Proceedings of the 15th Conference of the European Chapter of the Association for Computational Linguistics (EACL)},
 year   = {2017},
 pages   = {427--431}
}

@misc{BoersmaWeeninkPraat,
 author    = {Paul Boersma and David Weenink},
 title    = {Praat: Doing Phonetics by Computer [Computer program]},
 year     = {2025},
 howpublished = {\url{http://www.praat.org/}}
}

@misc{trusting_about,
  author       = {{TRUSTING Project}},
  title        = {About Us},
  howpublished = {\url{https://trusting-project.eu/about-us/}},
  note         = {Accessed: 2025-10-24},
  year         = {n.d.}
}

@article{huppi2025trusting,
  author       = {H{\"u}ppi, Roya M. and Bautista, Luc{\'i}a and Cecere, Giacomo and Just, Sandra A. and Koops, Sanne and Hussain, Musarrat and Tedeschi, Enrico and Bora, Emre and Lyne, John and Kaiser, Stefan and Spr{\"u}ngli-Toffel, Elodie and Kirschner, Matthias and Mikalsen, Karl {\O}yvind and Bongo, Lars Ailo and Van der Eycken, Erik and Spaniel, Filip and Elvev{\aa}g, Brita and Sommer, Iris E. and Hinzen, Wolfram and Homan, Philipp},
  title        = {TRUSTING: An International Multicenter Observational Study of Speech-Based Relapse Prediction in Psychosis Using Explainable AI},
  journal      = {medRxiv preprint},
  year         = {2025},
  doi          = {https://doi.org/10.1101/2025.11.14.25339774}
}

\section*{Author Contributions}

YP, FR, NL and PH conceptualized the project. YP wrote the codebase available on the GitHub repository. JBM reviewed the codebase. YP drafted the initial manuscript. YP, JBM, FR, VE, RH, SC, ARM, WH, IS, and PH critically revised and approved the final manuscript.

\end{document}